\documentclass[sigconf]{acmart}

\usepackage{amssymb}
\newcommand{\mathbold}[1]{\ensuremath{\boldsymbol{\mathbf{#1}}}}

\usepackage{lineno}

\usepackage{ragged2e}

\newcounter{parcount}

\DeclareRobustCommand{\parhead}[1]{\textbf{#1}~}

\definecolor{shadecolor}{gray}{0.9}

\usepackage{graphicx}
\usepackage[labelfont=bf]{caption}
\usepackage[format=hang]{subcaption}

\usepackage{booktabs, array, multirow}

\usepackage[algoruled]{algorithm2e}
\usepackage{algorithmic}
\usepackage{listings}
\usepackage{fancyvrb}
\fvset{fontsize=\small}

\usepackage[nameinlink]{cleveref}

\usepackage
[acronym,smallcaps,nowarn,section,nogroupskip,nonumberlist]{glossaries}
\glsdisablehyper{}

\lstdefinestyle{mystyle}{
    commentstyle=\color{OliveGreen},
    numberstyle=\tiny\color{black!60},
    stringstyle=\color{BrickRed},
    basicstyle=\ttfamily\scriptsize,
    breakatwhitespace=false,
    breaklines=true,
    captionpos=b,
    keepspaces=true,
    numbers=none,
    numbersep=5pt,
    showspaces=false,
    showstringspaces=false,
    showtabs=false,
    tabsize=2
}
\crefname{equation}{eq.}{eqs.}
\Crefname{equation}{Eq.}{Eqs.}
\creflabelformat{equation}{#1#2#3}

\usepackage{tikz}
\usetikzlibrary{bayesnet}
\usepackage[raggedright]{sidecap}
\sidecaptionvpos{figure}{t}

 \DeclareRobustCommand{\mb}[1]{\ensuremath{\boldsymbol{\mathbf{#1}}}}

\newcommand{\mbx}{\mathbold{x}}

\newcommand{\mbz}{\mathbold{z}}

\newcommand{\mbI}{\mathbold{I}}

\newcommand{\mbX}{\mathbold{X}}

\newcommand{\g}{\,\vert\,}

\newcommand{\EE}[2]{\mathbb{E}_{#1}\left[#2\right]}

\newcommand{\mult}{\mathrm{Mult}}

\newcommand{\cL}{\mathcal{L}}
\newcommand{\cN}{\mathcal{N}}

\DeclareMathOperator*{\argmax}{arg\,max}

\DeclareRobustCommand{\mb}[1]{\mathbold{#1}}

\newcommand{\mvae}{{\small Mult-}$\gls{VAE}^{\text{\sc pr}}$}
\newcommand{\mdae}{{\small Mult-}\gls{DAE}}

\newcommand{\vae}[1]{{\small #1-}$\gls{VAE}^{\text{\sc pr}}$}
\newcommand{\dae}[1]{{\small #1-}\gls{DAE}}
 \newacronym{ELBO}{elbo}{evidence lower bound}
\newacronym{VAE}{vae}{variational autoencoder}
\newacronym{WMF}{wmf}{weighted matrix factorization}
\newacronym{BPR}{bpr}{{B}ayesian personalized ranking}
\newacronym{CDAE}{cdae}{collaborative denoising autoencoder}
\newacronym{NCF}{ncf}{neural collaborative filtering}
\newacronym{DAE}{dae}{denoising autoencoder}
\newacronym{MLP}{mlp}{multilayer perceptron}

\begin{document}
\title{Variational Autoencoders for Collaborative Filtering}

\author{Dawen Liang}
\affiliation{  \institution{Netflix}
  \city{Los Gatos} 
  \state{CA} 
}
\email{dliang@netflix.com}

\author{Rahul G. Krishnan}
\affiliation{  \institution{MIT}
  \city{Cambridge} 
  \state{MA} 
}
\email{rahulgk@mit.edu}

\author{Matthew D. Hoffman}
\affiliation{  \institution{Google AI}
  \city{San Francisco} 
  \state{CA}}
\email{mhoffman@google.com}

\author{Tony Jebara}
\affiliation{  \institution{Netflix}
  \city{Los Gatos}
  \state{CA}}
\email{tjebara@netflix.com}

\renewcommand{\shortauthors}{D. Liang et al.}

\begin{abstract}

We extend \glspl{VAE} to collaborative filtering for implicit feedback. This 
non-linear probabilistic model enables us to go beyond the limited modeling capacity of 
linear factor models which still largely dominate collaborative filtering research.
We introduce a generative model with multinomial likelihood 
and use Bayesian inference for parameter estimation. 
Despite widespread use in language modeling and economics, the multinomial 
likelihood receives less attention in the recommender systems literature. 
We introduce a different 
regularization parameter for the learning objective, which proves to be crucial for achieving competitive performance.
Remarkably, there is an efficient way to tune the parameter using annealing. 
The resulting model and learning algorithm has information-theoretic connections to maximum entropy discrimination and the information bottleneck principle.
Empirically, we show that the proposed approach significantly outperforms several state-of-the-art baselines,
including two recently-proposed neural network approaches, on several real-world datasets.
We also provide extended experiments comparing the multinomial likelihood with other commonly 
used likelihood functions in the latent factor collaborative filtering literature and show 
favorable results. Finally, we identify the pros and cons of employing a principled Bayesian 
inference approach and characterize settings where it provides the most significant improvements.
\end{abstract}


\copyrightyear{2018}
\acmYear{2018} 
\setcopyright{iw3c2w3g}
\acmConference[WWW 2018]{The 2018 Web Conference}{April 23--27, 2018}{Lyon, France}
\acmPrice{}
\acmDOI{10.1145/3178876.3186150}
\acmISBN{978-1-4503-5639-8/18/04}

\fancyhead{}

\maketitle

\section{Introduction}
\glsreset{VAE}

Recommender systems are an integral component of the web. 
In a typical recommendation system, we observe how a set of users interacts with a set of items. Using this data, we seek to show users a set of previously unseen items they 
will like. As the web grows in size, good recommendation systems will play an important part in helping users interact more 
effectively with larger amounts of content. 

Collaborative filtering is among the most widely applied approaches in recommender systems. 
Collaborative filtering predicts what items a user will prefer by discovering and exploiting the similarity patterns across users and items. 
Latent factor models \citep{salakhutdinov2008probabilistic,hu2008collaborative,gopalan2015scalable} still 
largely dominate the collaborative filtering research literature due to their simplicity and effectiveness. 
However, these models are inherently linear, which limits their modeling capacity. Previous 
work \citep{liang2016factorization} has demonstrated that adding carefully crafted 
non-linear features into the linear latent factor models can significantly boost recommendation 
performance. Recently, a growing body of work involves applying neural networks to the collaborative filtering setting with promising results \citep{pmlr-v48-zheng16,sedhain2015autorec,wu2016collaborative,he2017neural}.  

Here, we extend \glspl{VAE} \citep{kingma2013auto,rezende2014stochastic}
to collaborative filtering for implicit feedback. \Glspl{VAE} generalize linear latent-factor models and enable us to
explore non-linear probabilistic latent-variable models, powered by neural networks, on large-scale recommendation datasets. 
We propose a neural generative model with multinomial conditional likelihood. Despite being widely used in language
modeling and economics \citep{blei2003latent,mcfadden1973conditional}, multinomial likelihoods appear
less studied in the collaborative filtering literature, particularly within the context of latent-factor models.
Recommender systems are often evaluated using ranking-based measures, such as mean 
average precision and normalized discounted cumulative gain \citep{jarvelin2002cumulated}. 
Top-$N$ ranking loss is difficult to optimize directly and previous work on direct ranking 
loss minimization resorts to relaxations and approximations \citep{weimer2008cofi,weston2011wsabie}.  
Here, we show that the multinomial likelihoods are well-suited for modeling implicit feedback data, and are a closer proxy to the ranking 
loss relative to more popular likelihood functions such as Gaussian and logistic. 

Though recommendation is often considered a big-data problem (due to the huge numbers of users
and items typically present in a recommender system), we argue that, in contrast, it represents a uniquely 
challenging ``small-data'' problem: most users only interact with a tiny proportion of the items 
and our goal is to collectively make informed inference about each user's preference. To make use of the sparse 
signals from users and avoid overfitting, we build a probabilistic latent-variable model that shares 
statistical strength among users and items. Empirically, we show that employing a 
principled Bayesian approach is more robust regardless of the scarcity of the data.

Although \glspl{VAE} have been extensively studied for image modeling and generation,
there is surprisingly little work applying \glspl{VAE} to recommender systems.
We find that two adjustments are essential to getting state-of-the-art results with \glspl{VAE} on this task:
\begin{itemize}
\item First, we use a multinomial likelihood for the data distribution. We show that this simple choice
realizes models that outperform the more commonly used Gaussian and logistic likelihoods.
\item Second, we reinterpret and adjust the standard \gls{VAE} objective, which we argue is over-regularized.
We draw connections between the learning algorithm resulting from our proposed regularization 
and the information-bottleneck principle and maximum-entropy discrimination. 
\end{itemize}
The result is a recipe that makes \glspl{VAE} practical solutions to this important problem.
Empirically, our methods significantly outperform 
state-of-the-art baselines on several real-world datasets, including
two recently proposed neural-network approaches.
 
\section{Method}\label{sec:method}

We use $u \in \{1, \dots, U\}$ to index users and $i \in \{1, \dots, I\}$ to index
items. In this work, we consider learning with implicit feedback \citep{hu2008collaborative,pan2008one}. 
The user-by-item interaction matrix is the
click\footnote{We use the verb ``click'' for concreteness; this can be any type of interaction, 
including ``watch'', ``purchase'', or ``listen''.} matrix $\mbX \in \mathbb{N}^{U \times I}$. 
The lower case $\mbx_u = [x_{u1}, \dots, x_{uI}]^\top \in \mathbb{N}^I$ is a bag-of-words vector with the number 
of clicks for each item from user $u$. For simplicity, we binarize the click matrix. 
It is straightforward to extend it to general count data. 

\subsection{Model}\label{sec:model}

The generative process we consider in this paper is similar to 
the deep latent Gaussian model \citep{rezende2014stochastic}. For each user $u$, the model starts by 
sampling a $K$-dimensional latent representation $\mbz_u$ from a standard Gaussian prior. 
The latent representation $\mbz_u$ is transformed via a 
non-linear function 
$f_\theta(\cdot) \in \mathbb{R}^I$ 
to produce a probability distribution over $I$ items $\pi(\mbz_u)$ from which 
the click history $\mbx_u$ is assumed to have been drawn:
\begin{equation}
\begin{split}
\mbz_u \sim \cN(&0, \mbI_K), \quad\pi(\mbz_u) \propto \exp\{f_\theta(\mbz_u)\},\\
&\mbx_u \sim \mult(N_u, \pi(\mbz_u)).
\end{split}
\label{eq:gen_model}
\end{equation}
The non-linear function $f_\theta(\cdot)$ is a multilayer perceptron with parameters $\theta$.  
The output of this transformation is normalized via a softmax function to 
produce a probability vector $\pi(\mbz_u) \in \mathbb{S}^{I-1}$ (an $(I-1)$-simplex) 
over the entire item set. 
Given the total number of clicks $N_u = \sum_i x_{ui}$ from user $u$, the observed bag-of-words vector $\mbx_u$ is assumed to be 
sampled from a multinomial distribution with probability $\pi(\mbz_u)$. 
This generative model generalizes the 
latent-factor model --- we can recover classical matrix factorization \citep{salakhutdinov2008probabilistic} by
setting $f_\theta(\cdot)$ to be linear and using a Gaussian likelihood. 

The log-likelihood for user $u$ (conditioned on the latent representation) is: 
\begin{equation}\label{eq:multi}
\log p_\theta(\mbx_u \g \mbz_u) \overset{c}{=} \sum_i x_{ui} \log \pi_i(\mbz_u).
\end{equation}
This multinomial likelihood is commonly used in language models, e.g., latent Dirichlet 
allocation \citep{blei2003latent}, and economics, e.g., multinomial logit choice model \citep{mcfadden1973conditional}. 
It is also used in the cross-entropy loss\footnote{The cross-entropy loss for multi-class classification 
is a multinomial likelihood under a single draw from the distribution.} for multi-class classification. For example, it has been used in recurrent neural networks for 
session-based sequential recommendation \citep{hidasi2015session,tan2016improved,smirnova2017contextual,hidasi2017recurrent,chatzis2017recurrent} and 
in feedward neural networks applied to Youtube recommendation \citep{covington2016deep}. 
The multinomial likelihood is less well studied in the context of latent-factor models such as matrix factorization and autoencoders.
A notable exception is the collaborative competitive filtering (CCF) model \citep{yang2011collaborative} and its successors,
which take advantage of more fine-grained information about what options were presented to which users.
(If such information is available, it can also be incorporated into our \gls{VAE}-based approach.)

We believe the multinomial distribution is well suited to modeling click data.
The likelihood of the click matrix (\Cref{eq:multi}) rewards the model for putting probability mass on the non-zero entries in $\mbx_u$.
But the model has a limited budget of probability mass, since $\pi(\mbz_u)$ must sum to 1;
the items must compete for this limited budget \citep{yang2011collaborative}.
The model should therefore assign more probability mass to items that are more likely to be clicked.
To the extent that it can, it will perform well under the top-$N$ ranking loss that recommender systems are commonly evaluated on.

By way of comparison, we present two popular choices of likelihood functions used in latent-factor collaborative filtering: Gaussian and logistic likelihoods.
Define $f_\theta(\mbz_u) \equiv [f_{u1}, \dots, f_{uI}]^\top$ as the output of the generative function $f_\theta(\cdot)$. The Gaussian log-likelihood for user $u$ is
\begin{equation}\label{eq:gaussian}
\log p_\theta(\mbx_u \g \mbz_u) \overset{c}{=} -\sum_i \frac{c_{ui}}{2} (x_{ui} - f_{ui})^2.
\end{equation}
We adopt the convention in \citet{hu2008collaborative} and introduce a ``confidence'' weight $c_{x_{ui}} \equiv c_{ui}$ where $c_1 > c_0$ to balance the unobserved 0's which far outnumber the observed 1's in most click data. This is also equivalent to training the model with unweighted Gaussian likelihood and negative sampling. The logistic log-likelihood\footnote{Logistic likelihood is also cross-entropy loss for binary classification.} for user $u$ is 
\begin{equation}\label{eq:logistic}
\log p_\theta(\mbx_u \g \mbz_u) = \sum_i x_{ui} \log \sigma(f_{ui}) + (1 - x_{ui}) \log (1 - \sigma(f_{ui})),
\end{equation}
where $\sigma(x) = 1/(1 + \exp(-x))$ is the logistic function. We compare multinomial likelihood with Gaussian and logistic in \Cref{sec:exp}.

\subsection{Variational inference} \label{sec:vi}

To learn the generative model in \Cref{eq:gen_model}, we are interested in estimating 
$\theta$ (the parameters of $f_\theta(\cdot)$). To do so, for each data point we need to approximate the intractable posterior distribution $p(\mbz_{u} \g \mbx_{u})$.
We resort to variational inference \citep{jordan1999introduction}.
Variational inference approximates the true intractable posterior with a simpler variational distribution $q(\mbz_u)$. 
We set $q(\mbz_u)$ to be a fully factorized (diagonal) Gaussian distribution:
\[q(\mbz_u) = \cN(\mb\mu_u, \mathrm{diag}\{\mb\sigma_u^2\}).\]
The objective of variational inference is to optimize the free variational parameters $\{\mb\mu_u, \mb\sigma_u^2\}$ so that the Kullback-Leiber divergence $\mathrm{KL}(q(\mbz_u) \| p(\mbz_u | \mbx_u ))$ is minimized.

\glsreset{VAE}
\subsubsection{Amortized inference and the variational autoencoder:} 
With variational inference the number of parameters to optimize $\{\mb\mu_u, \mb\sigma_u^2\}$ grows with the 
number of users and items in the dataset. This can become a bottleneck for commercial 
recommender systems with millions of users and items. The \gls{VAE} \citep{kingma2013auto,rezende2014stochastic} replaces individual variational 
parameters with a data-dependent function (commonly called an \emph{inference model}):
\[
g_\phi(\mbx_u) \equiv [\mu_\phi(\mbx_u), \sigma_\phi(\mbx_u)] \in \mathbb{R}^{2K} 
\]
parametrized by $\phi$ with both $\mu_\phi(\mbx_u)$ and $\sigma_\phi(\mbx_u)$ being $K$-vectors and sets the variational distribution as follows:
\[
q_\phi(\mbz_u \g \mbx_u) = \cN(\mu_\phi(\mbx_u), \mathrm{diag}\{\sigma_\phi^2(\mbx_u)\}).
\]
That is, using the observed data $\mbx_u$ as input, the inference model outputs the corresponding 
variational parameters of variational distribution $q_\phi(\mbz_u \g \mbx_u)$, which, when 
optimized, approximates the intractable posterior $p(\mbz_u \g \mbx_u)$.\footnote{In the implementation, the inference 
model will output the log of the variance of the variational distribution. 
We continue to use $\sigma_\phi(\mbx_u)$ for notational brevity.} 
Putting $q_\phi(\mbz_u \g \mbx_u)$ and the generative model $p_\theta(\mbx_u \g \mbz_u)$ together in \Cref{fig:autoencoders}c, we end up 
with a neural architecture that resembles an autoencoder --- hence the name variational autoencoder. 

\Glspl{VAE} make use of amortized inference \citep{gershman2014amortized}: they flexibly reuse inferences
to answer related new problems. This is well aligned with the ethos of collaborative 
filtering: analyze user preferences by exploiting the similarity patterns inferred from 
past experiences. In \Cref{sec:pred}, we discuss how this enables us to perform prediction efficiently.

\parhead{Learning \glspl{VAE}: } 
As is standard when learning latent-variable models with variational inference \citep{doi:10.1080/01621459.2017.1285773}, we can
lower-bound the log marginal likelihood of the data. This forms the objective we seek to maximize 
for user $u$ (the objective function of the dataset is obtained by averaging the objective function over all the users):
\begin{equation}
\begin{split}
\log p(\mbx_u; \theta) &\geq \EE{q_\phi(\mbz_u \g \mbx_u)}{\log p_\theta(\mbx_u \g \mbz_u)} - \mathrm{KL}(q_\phi(\mbz_u \g \mbx_u) \| p(\mbz_u))\\
&\equiv \cL(\mbx_u; \theta, \phi)
\end{split}\label{eq:elbo}
\end{equation}
This is commonly known as the \gls{ELBO}. Note that the \gls{ELBO} is a function of both $\theta$ and $\phi$. We can obtain an unbiased estimate of \gls{ELBO} by sampling $\mbz_u \sim q_\phi$ and perform stochastic gradient ascent to optimize it. However, the challenge is that we cannot trivially take gradients with respect to $\phi$ through this sampling process. The \emph{reparametrization trick} \citep{kingma2013auto,rezende2014stochastic} sidesteps this issue: we sample $\mb\epsilon\sim\cN(0, \mbI_K)$ and reparametrize $\mbz_u = \mu_\phi(\mbx_u)+ \mb\epsilon \odot \sigma_\phi(\mbx_u)$. By doing so, the stochasticity in the sampling process is isolated and the gradient with respect to $\phi$ can be back-propagated through the sampled $\mbz_u$. The \gls{VAE} training procedure is summarized in \Cref{alg:vae}.

\begin{algorithm}
\DontPrintSemicolon \KwIn{Click matrix $\mbX \in \mathbb{R}^{U \times I}$}
Randomly initialize $\theta$, $\phi$\;
\While{not converged}{
  Sample a batch of users $\mathcal{U}$\;
  \ForAll{$u\in\mathcal{U}$}{
    Sample $\mb\epsilon\sim \cN(0, \mbI_K)$ and compute $\mbz_u$ via reparametrization trick\;
    Compute noisy gradient $\nabla_\theta \cL$ and $\nabla_\phi \cL$ with $\mbz_u$\;
  }
  Average noisy gradients from batch\;
  Update $\theta$ and $\phi$ by taking stochastic gradient steps\;
}
\Return{$\theta$, $\phi$}\;
\caption{{\sc VAE-SGD} Training collaborative filtering \gls{VAE} with stochastic gradient descent.}
\label{alg:vae}
\end{algorithm}

\subsubsection{Alternative interpretation of \gls{ELBO}.} \label{sec:reg}
We can view \gls{ELBO} defined in \Cref{eq:elbo} from a different perspective: the first term can be interpreted as (negative) reconstruction error, while 
the second KL term can be viewed as regularization. It is this perspective we work with because it allows us to make a trade-off that forms the crux of our method.  
From this perspective, it is natural to extend the \gls{ELBO} by introducing a parameter $\beta$ to control the strength of regularization:
\begin{equation} \label{eq:elbo_anneal}
\begin{split}
\cL_\beta(\mbx_u; \theta, \phi) \equiv \mathbb{E}_{q_\phi(\mbz_u \g \mbx_u)}[&\log p_\theta(\mbx_u \g \mbz_u)] \\
-& \beta\cdot \mathrm{KL}(q_\phi(\mbz_u \g \mbx_u) \| p(\mbz_u)).
\end{split}
\end{equation}

While the original \gls{VAE} (trained with \gls{ELBO} in \Cref{eq:elbo}) is a powerful generative model;
we might ask whether
we need \emph{all} the statistical properties of a generative model for tackling problems in recommender systems. 
In particular, if we are willing to sacrifice the ability to perform ancestral sampling, can we improve our performance?
The regularization view of the \gls{ELBO} (\Cref{eq:elbo_anneal}) introduces a trade-off between how well we can fit the data and 
how close the approximate posterior stays to the prior during learning.

We propose using $\beta \neq 1$. This means we are no longer optimizing a lower bound on the log marginal likelihood.
If $\beta < 1$, then we are also weakening the influence of the prior constraint
$\frac{1}{U}\sum_u q(\mbz \g \mbx_u)\approx p(\mbz) = \mathcal{N}(\mbz; 0, \mbI_K)$
\citep{hoffman2016elbo}; this means that the model is less able to generate novel user histories by ancestral sampling.

But ultimately our goal is to make good recommendations, not to maximize likelihood or generate imagined user histories.
Treating $\beta$ as a free regularization parameter therefore costs us nothing, and, as we will see, yields significant improvements in performance.

\parhead{Selecting $\beta$:} 
We propose a simple heuristic for setting $\beta$: we start
training with $\beta = 0$, and gradually increase $\beta$ to 1. 
We linearly anneal the KL term slowly over a large number of gradient updates to $\theta,\phi$ and record the best $\beta$ when its performance reaches the peak.
We found this method to work well and it does not require the need for training multiple models with different values of $\beta$, which can be time-consuming. 
Our procedure is inspired by KL annealing \citep{bowman2015generating}, a common heuristic used for training \glspl{VAE} when there is concern that the model is being underutilized. 

\begin{figure}[!ht]
  \centering
  \includegraphics[width=\columnwidth]{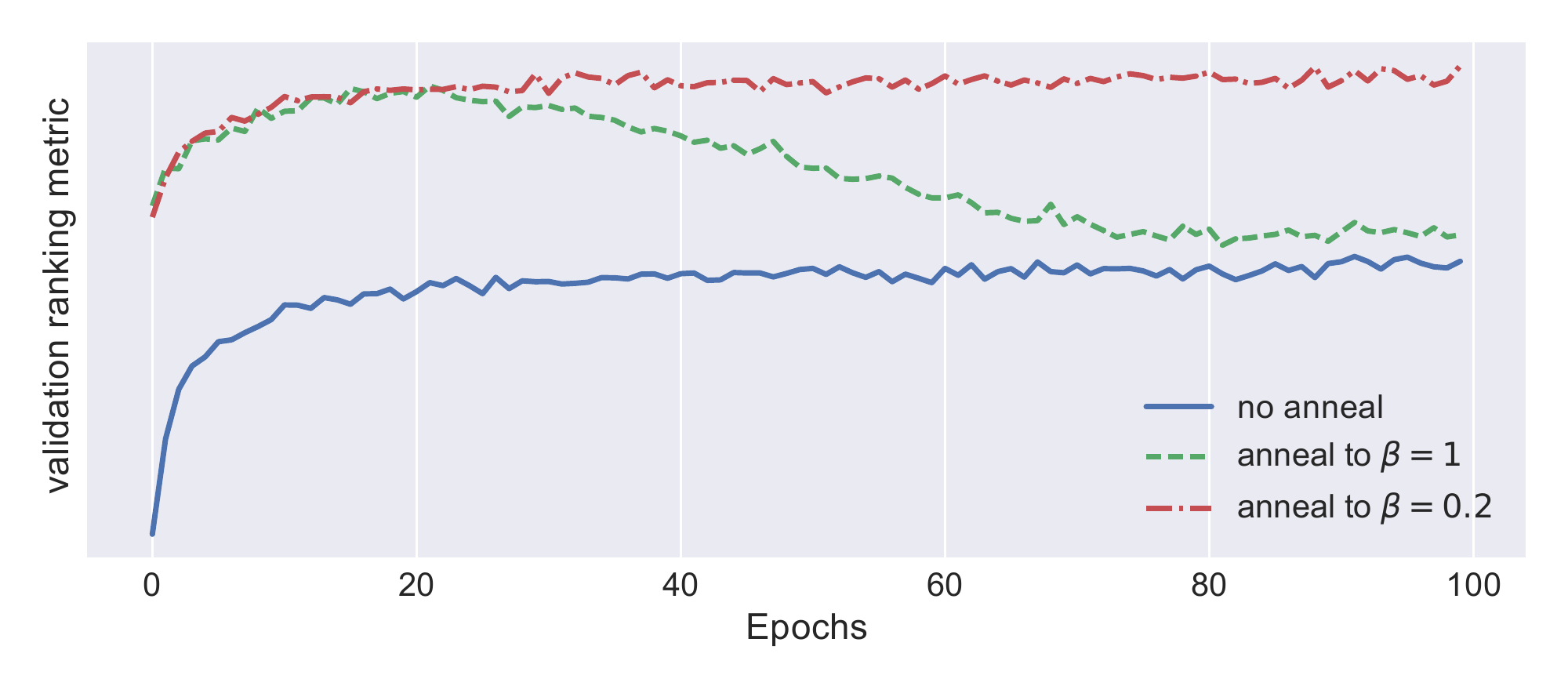}
  \caption{Validation ranking metrics with different annealing configurations. For the green dashed curve, $\beta$ reaches 1 at around 80 epochs.}
  \label{fig:kl_anneal}
\end{figure}

\Cref{fig:kl_anneal} illustrates the basic idea (we observe the same trend consistently across datasets). 
Here we plot the validation ranking metric without KL annealing (blue solid) 
and with KL annealing all the way to $\beta = 1$ (green dashed, $\beta$ reaches 1 at around 80 epochs). 
As we can see, the performance is poor without any KL annealing. 
With annealing, the validation performance first increases as the training proceeds and then drops as $\beta$ gets close to 1 to a value that is only slightly better
than doing no annealing at all. 

Having identified the best $\beta$ based on the peak validation metric, we can 
retrain the model with the same annealing schedule, but stop increasing $\beta$ after reaching 
that value (shown as red dot-dashed in \Cref{fig:kl_anneal}).\footnote{We found this to give slightly better results 
than keeping $\beta$ at the best value throughout the training.} This might be sub-optimal 
compared to a thorough grid search. However, it is much more efficient, and gives us competitive empirical performance. If the computational budget 
is scarce, then within a single run, we can stop increasing $\beta$ when we notice the validation metric dropping.
Such a procedure incurs no additional runtime to learning a standard \gls{VAE}.
We denote this partially regularized \gls{VAE} with multinomial likelihood as \mvae.

\subsubsection{Computational Burden} Previous collaborative filtering models with 
neural networks \citep{he2017neural,wu2016collaborative} are trained with stochastic gradient descent
where in each step a single (user, item) entry from the click matrix is randomly sampled to perform a gradient update. 
In \Cref{alg:vae} we subsample users and take their entire click history (complete rows of the click matrix) to update model parameters.
This eliminates the necessity of negative sampling (and consequently the hyperparameter tuning for picking the number of negative examples), commonly used in the (user, item) entry subsampling scheme. 

A computational challenge that comes with our approach, however, is that when the number of items is huge, computing the 
multinomial probability $\pi(\mbz_u)$ could be computationally expensive, since it requires computing the predictions 
for all the items for normalization. This is a common challenge for language modeling where the size of 
the vocabulary is in the order of millions or more \citep{mikolov2013distributed}. In our experiments on some medium-to-large 
datasets with less than 50K items (\Cref{sec:exp_data}), this has not yet come up as a 
computational bottleneck. 
If this becomes a bottleneck when working with larger item sets, one can easily apply the
simple and effective method proposed by \citet{pmlr-v54-botev17a} to approximate the normalization 
factor for $\pi(\mbz_u)$.

\subsection{A taxonomy of autoencoders} \label{sec:mle}

In \Cref{sec:vi}, we introduced maximum marginal likelihood estimation of \glspl{VAE} using 
approximate Bayesian inference under a non-linear generative model (\Cref{eq:gen_model}). 
We now describe our work from the perspective of learning autoencoders.  
Maximum-likelihood estimation in a regular autoencoder takes the following form:
\begin{equation}\label{eq:mle}
\begin{split}
\theta^{\mathrm{AE}}, \phi^{\mathrm{AE}} =&\argmax_{\theta, \phi} \sum_u \EE{\delta(\mbz_u - g_\phi(\mbx_u))}{\log p_\theta(\mbx_u \g \mbz_u)}\\
= &\argmax_{\theta, \phi} \sum_u \log p_\theta(\mbx_u \g g_\phi(\mbx_u))
\end{split}
\end{equation}
There are two key distinctions of note: 
(1) The autoencoder (and denoising autoencoder) effectively optimizes the first term in the \gls{VAE} objective (\Cref{eq:elbo} 
and \Cref{eq:elbo_anneal}) using a delta variational distribution $q_\phi(\mbz_u \g \mbx_u) = \delta(\mbz_u - g_\phi(\mbx_u))$ --- 
it does not regularize $q_\phi(\mbz_u \g\mbx_u)$ towards any prior distribution as the \gls{VAE} does. 
(2) the $\delta(\mbz_u - g_\phi(\mbx_u))$ is a $\delta$ distribution with 
mass only at the output of $g_{\phi}(\mbx_u)$. Contrast this to the \gls{VAE}, where the learning is done using a variational \emph{distribution},
i.e., $g_{\phi}(\mbx_u)$ outputs the parameters (mean and variance) of a Gaussian distribution. This means that \gls{VAE} has the ability to capture
per-data-point variances in the latent state $\mbz_u$. 

In practice, we find that learning autoencoders is  
extremely prone to overfitting as the network learns to put all the probability mass to the non-zero
entries in $\mbx_u$. By introducing dropout \citep{srivastava2014dropout} at the input layer, the \gls{DAE} is less
prone to overfitting and we find that it also gives competitive empirical results. In addition to the \mvae, we also study a denoising autoencoder 
with a multinomial likelihood. We denote this model \mdae. 
In \Cref{sec:exp} 
we characterize the tradeoffs in what is gained and lost by explicitly 
parameterizing the per-user variance with \mvae~versus using a point-estimation in \mdae.

\begin{figure*}[h]
\centering
	\begin{tikzpicture}[scale=0.75, transform shape,blackdot/.style={thin, draw=black, align=center, scale = 0.3,fill=black}]
	\node [latent] (z_ae) {$z$};
	\node [obs,below= of z_ae] (x_ae) {$x$};
	\node [obs,above= of z_ae] (xq_ae) {$x$};
	\draw [->] (z_ae) to node[pos=0.45,fill=white] {$\theta$} (x_ae);
	\draw [->] (xq_ae) to node[pos=0.45,fill=white] {$\phi$} (z_ae);

	\node [obs,right= of xq_ae,xshift=3cm] (xq_dae) {$x$};
	\node [latent,right= of z_ae,xshift=3cm] (z_dae) {$z$};
	\node [obs,right= of z_dae,yshift=1cm] (eps_dae) {$\epsilon$};
	\node [obs,below= of z_dae] (x_dae) {$x$};
	\draw [->] (z_dae) to node[pos=0.45,fill=white] {$\theta$} (x_dae);
	\draw [->] (xq_dae) to node[pos=0.45,fill=white] {$\phi$} (z_dae);
	\draw [->,dashed] (eps_dae) -- (xq_dae);

	\node [obs,right= of xq_dae,xshift=5cm] (xq_vae) {$x$};
	\node [latent,below= of xq_vae,xshift=-1.35cm,yshift=0.8cm] (mu_vae) {$\mu$};
	\node [latent,below= of xq_vae,xshift=1.35cm,yshift=0.8cm] (sigma_vae) {$\sigma$};
	\node [latent,right= of z_dae,xshift=5cm] (z_vae) {$z$};
	\node [obs,right= of z_vae] (eps_vae) {$\epsilon$};
	\node [obs,below= of z_vae] (x_vae) {$x$};
	\draw [->] (z_vae) to node[pos=0.4,fill=white] {$\theta$} (x_vae);
	\draw [->] (xq_vae) to node[pos=0.35,fill=white] {$\phi$} (mu_vae);
	\draw [->] (xq_vae) to node[pos=0.35,fill=white] {$\phi$} (sigma_vae);
	\draw [->] (sigma_vae) -- (z_vae);
	\draw [->] (mu_vae) -- (z_vae);
	\draw [->,dashed] (eps_vae) -- (z_vae);

	\node[text width=3cm, anchor=west, right] at (-3,0)
    {(a) Autoencoder};

	\node[text width=4cm, anchor=west, right] at (2,0)
    {(b) Denoising \newline Autoencoder};

	\node[text width=4cm, anchor=west, right] at (8,0)
    {(c) Variational \newline Autoencoder};
	\end{tikzpicture}
	\caption{
	\small
	A taxonomy of autoencoders. The dotted arrows denote a sampling operation.
	}
	\label{fig:autoencoders}
\end{figure*}
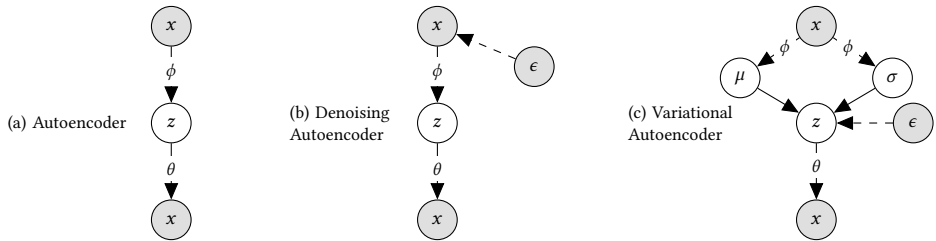
 To provide a unified view of different variants of autoencoders and clarify where our work stands,
we depict variants of autoencoders commonly found in the literature in \Cref{fig:autoencoders}. 
For each one, we specify the model (dotted arrows denote a sampling operation) and describe the training objective used in parameter estimation.

In \Cref{fig:autoencoders}a we have autoencoder. It is trained to reconstruct input with the same objective as in \Cref{eq:mle}. 
Adding noise to the input (or the intermediate hidden representation) 
of an autoencoder yields the denoising autoencoder in \Cref{fig:autoencoders}b. The training objective is the same as that of an autoencoder. 
\mdae~belongs to this model class. \Glsfirst{CDAE} \citep{wu2016collaborative} is a variant of this model class. 
The \gls{VAE} is depicted in \Cref{fig:autoencoders}c. Rather than using a delta variational distribution, it uses an inference model parametrized by 
$\phi$ to produce the mean and variance of the approximating variational distribution. 
The training objective of the \gls{VAE} is given in \Cref{eq:elbo_anneal}. 
Setting $\beta$ to 1 recovers the original \gls{VAE} formulation \citep{kingma2013auto,rezende2014stochastic}. 
\citet{higgins2017beta} study the case where $\beta > 1$. Our model, \mvae~ corresponds to learning \glspl{VAE} with $\beta\in[0,1]$.

\subsection{Prediction}\label{sec:pred}

We now describe how we make predictions given a trained generative model of the form 
\Cref{eq:gen_model}. For both, \mvae~ (\Cref{sec:vi}) or \mdae~ (\Cref{sec:mle}), we make predictions in the same way. 
Given a user's click history $\mbx$, we rank all the items based on the un-normalized predicted multinomial probability $f_\theta(\mbz)$. 
The latent representation $\mbz$ for $\mbx$ is constructed as follows: 
For \mvae, we simply take the mean of the variational distribution $\mbz = \mu_\phi(\mbx)$; for \mdae, we take the output $\mbz = g_\phi(x)$. 

It is easy to see the advantage of using autoencoders. 
We can effectively make predictions for users by evaluating two functions -- the inference model (encoder) $g_\phi(\cdot)$ and the generative model (decoder) $f_\theta(\cdot)$. 
For most of the latent factor collaborative filtering model, e.g., matrix factorization 
\citep{hu2008collaborative,gopalan2015scalable}, when given the click history of a user that is not 
present in the training data, normally we need to perform some form of optimization to obtain the latent factor 
for this user. 
This makes the use of autoencoders particularly attractive in industrial applications,
where it is important that predictions be made cheaply and with low latency.
 
\section{Related work} \label{sec:related}
\parhead{\Glspl{VAE} on sparse data.}
\glsreset{VAE} \Glspl{VAE} \citep{kingma2013auto,rezende2014stochastic} have seen much application to images since their inception. 
\citet{doersch2016tutorial} presents a review on different applications of \gls{VAE} to image data. \citet{miao2016neural} study \gls{VAE}s on text data.
More recent results from \citet{krishnan2017challenge} find that \glspl{VAE} (trained with \Cref{eq:elbo}) suffer from underfitting when modeling 
large, sparse, high-dimensional data. We notice similar issues when fitting \gls{VAE} without annealing (\Cref{fig:kl_anneal}) or annealing to $\beta=1$.  
By giving up the ability to perform ancestral sampling in the model, and setting $\beta \leq 1$, the resulting model is 
no longer a proper \emph{generative} model though for collaborative filtering tasks we always make predictions \emph{conditional on} users' click history.

\parhead{Information-theoretic connection with \gls{VAE}.} 
The regularization view of the \gls{ELBO} in \Cref{eq:elbo_anneal} 
resembles maximum-entropy discrimination \citep{jaakkola2000maximum}. 
Maximum-entropy discrimination attempts to combine discriminative estimation with 
Bayesian inference and generative modeling. In our case, in \Cref{eq:elbo_anneal}, $\beta$ acts as a knob to balance 
discriminative and generative aspects of the model. 

The procedure in \Cref{eq:elbo_anneal} has information-theoretic connections 
described in \citet{alemi2016deep}. The authors propose the deep variational information bottleneck, which is a variational approximation 
to the information 
bottleneck principle \citep{tishby2000information}. They show that as a special case they can recover the learning objective used
by \glspl{VAE}. They report more robust supervised classification performance 
with $\beta < 1$. This is consistent with our findings as well. 
\citet{higgins2017beta} proposed $\beta$-\gls{VAE}, which leads 
to the same objective as \Cref{eq:elbo_anneal}. They motivate $\beta$-\gls{VAE} for the goal of learning disentangled representations
from images (basic visual concepts, such as shape, scale, and color). Their work, however, sets $\beta \gg 1$, effectively imposing 
a stronger independent prior assumption on the latent code $\mbz$. While their motivations are quite different from ours, it is 
interesting to note orthogonal lines of research emerging from exploring the full spectrum of values for $\beta$. 

\parhead{Neural networks for collaborative filtering.} 
Early work on neural-network-based collaborative filtering models focus on explicit 
feedback data and evaluates on the task of 
rating predictions \citep{salakhutdinov2007restricted,pmlr-v28-georgiev13,sedhain2015autorec,pmlr-v48-zheng16}. 
The importance of implicit feedback has been gradually recognized, and consequently most recent research, such as this work, has focused on it. 
The two papers that are most closely related 
to our approaches are \glsfirst{CDAE} \citep{wu2016collaborative} and \glsfirst{NCF} \citep{he2017neural}. 

\Gls{CDAE} \citep{wu2016collaborative} augments the standard denoising autoencoder, described in \Cref{sec:mle}, by adding a 
per-user latent factor to the input. The number of parameters of the \gls{CDAE} model grows linearly with both 
the number of users as well as the number of items, making it more prone to overfitting.
In contrast, the number of parameters in the \gls{VAE} grows linearly with the number of items. The \gls{CDAE} also requires additional optimization to obtain the latent factor for unseen users to make predicion. 
In the paper, the authors investigate the Gaussian and logistic likelihood 
loss functions --- as we show, the multinomial likelihood is significantly more robust for use in recommender systems. 
\Gls{NCF} \citep{he2017neural} explore a model with non-linear interactions between the user and item latent 
factors rather than the commonly used dot product. The authors demonstrate improvements of \gls{NCF} over standard 
baselines on two small datasets. Similar to \gls{CDAE}, the number of parameters of \gls{NCF} 
also grows linearly with both the number of users as well as items. We find that this becomes problematic for much 
larger datasets. We compare with both \gls{CDAE} and \gls{NCF} in \Cref{sec:exp}.

Asymmetric matrix factorization \citep{paterek2007improving} may also be interpreted as an autoencoder, as 
elaborated in \citet{steck2015gaussian}. We can recover this work by setting both $f_\theta(\cdot)$ and $g_\phi(\cdot)$ to be linear.

Besides being applied in session-based sequential recommendation (see \Cref{sec:model}),
various approaches \citep{NIPS2013_5004,liang2015content,almahairi2015learning,wang2015collaborative} have applied neural networks 
to incorporate \emph{side information} into collaborative filtering models to better handle the cold-start problem.
These approaches are complementary to ours.
 
\section{Empirical Study} \label{sec:exp}

We evaluate the performance of \mvae~and \mdae. We provide insights into their performance by exploring the resulting fits. We highlight the following results:  
\begin{itemize}
	\item \mvae~achieves state-of-the-art results on three real-world datasets when compared with various baselines, including recently proposed neural-network-based collaborative filtering models. 
	\item For the denoising and variational autoencoder, the multinomial likelihood compares favorably over the more common Gaussian and logistic likelihoods. 
	\item Both \mvae~ and \mdae~ produce competitive empirical results. We identify when parameterizing the uncertainty explicitly as in \mvae~ does better/worse than the point estimate used by \mdae~ and list pros and cons for both approaches.
\end{itemize}

The source code to reproduce the experimental results is available on GitHub\footnote{\url{https://github.com/dawenl/vae_cf}}. 

\subsection{Datasets} \label{sec:exp_data}

We study three medium- to large-scale user-item consumption
datasets from various domains: 

\parhead{MovieLens-20M (ML-20M):} These are user-movie ratings collected
from a movie recommendation service. We binarize the
explicit data by keeping ratings of four or higher and
interpret them as implicit feedback. We only keep users who
have watched at least five movies.

\parhead{Netflix Prize (Netflix):} This is the user-movie ratings data from the Netflix Prize\footnote{\url{http://www.netflixprize.com/}}. Similar to ML-20M, we binarize explicit data by keeping ratings of four or higher. We only keep users who have watched at least five movies. 

\parhead{Million Song Dataset (MSD):} This data contains the user-song play counts released as part of the Million Song Dataset \citep{bertin2011million}. We binarize play
counts and interpret them as implicit preference data. We only
keep users with at least 20 songs in their listening history
and songs that are listened to by at least 200 users.

\Cref{tab:data} summarizes the dimensions
of all the datasets after preprocessing.

\begin{table}
\begin{center}
\caption{Attributes of datasets after preprocessing. Interactions
are non-zero entries. \%  of interactions refers to the density of the user-item
click matrix $\mbX$. \# of the held-out users is the number of validation/test users out of the total number of users in the first row. }
\begin{tabular}{ l c c c c }
  \toprule
   & \textbf{ML-20M} & \textbf{Netflix} & \textbf{MSD}  \\
  \midrule
  \# of users & 136,677 & 463,435 & 571,355  \\
  \# of items & 20,108 & 17,769 & 41,140  \\
  \# of interactions & 10.0M & 56.9M & 33.6M\\
  \% of interactions & 0.36\% & 0.69\% & 0.14\% \\
  \midrule
  \# of held-out users & 10,000 & 40,000 & 50,000 \\
  \bottomrule
\end{tabular}
\label{tab:data}
\end{center}
\end{table}

\subsection{Metrics}\label{sec:exp_metrics}

We use two ranking-based metrics: Recall@$R$ and the truncated normalized
discounted cumulative gain (NDCG@$R$). For each user, both metrics compare the predicted rank of the held-out items with their
true rank. For both \mvae~ and \mdae, we get the predicted rank by sorting the un-normalized multinomial probability $f_\theta(\mbz)$. While Recall@$R$ considers all items ranked within
the first $R$ to be equally important, NDCG@$R$ uses
a monotonically increasing discount to emphasize the importance
of higher ranks versus lower ones. Formally, define $\omega(r)$ as the item at rank $r$, $\mathbb{I}[\cdot]$ is the indicator function, and $I_u$ is the set of held-out items that user $u$ clicked on.

Recall@$R$ for user $u$ is
\[
\textrm{Recall@}R(u, \omega) :=  \frac{\sum_{r=1}^R \mathbb{I}[\omega(r) \in I_u]}{\min(M, |I_u|)}.
\]

The expression in the denominator is the minimum
of $R$ and the number of items clicked on by user
u. This normalizes Recall@$R$ to have a maximum of 1,
which corresponds to ranking all relevant items in the top
$R$ positions.

Truncated discounted cumulative gain (DCG@$R$) is
\[
\textrm{DCG@}R(u, \omega) := \sum_{r=1}^R \frac{2^{\mathbb{I}[\omega(r) \in I_u]} - 1}{\log(r+1)}.
\]

NDCG@$R$ is the DCG@$R$ linearly normalized to [0, 1] after dividing by the best possible DCG@$R$, where all the held-out items are ranked at the top. 

\subsection{Experimental setup}
We study the performance of various models under strong generalization \citep{marlin2004collaborative}: We split all users into training/validation/test sets. We train models using the entire click history of the training users. To evaluate, we take part of the click history from held-out (validation and test) users to learn the necessary user-level representations for the model and then compute metrics by looking at how well the model ranks the rest of the unseen click history from the held-out users. 

This is relatively more difficult than weak generalization where the user's click history can appear during both training and evaluation. We consider it more realistic and robust as well. In the last row of \Cref{tab:data}, we list the number of held-out users (we use the same number of users for validation and test). For each held-out user, we randomly choose 80\% of the click history as the ``fold-in'' set to learn the necessary user-level representation and report metrics on the remaining 20\% of the click history. 

We select model hyperparameters and architectures by evaluating NDCG@100 on the validation users. For both \mvae~ and \mdae, we keep the architecture for the generative model $f_\theta(\cdot)$ and the inference model $g_\phi(\cdot)$ symmetrical and explore \gls{MLP} with 0, 1, and 2 hidden layers. We set the dimension of the latent representation $K$ to 200 and any hidden layer to 600. As a concrete example, recall $I$ is the total number of items, the overall architecture for a \mvae/\mdae~ with 1-hidden-layer \gls{MLP} generative model would be $[I \rightarrow 600 \rightarrow 200 \rightarrow 600 \rightarrow I]$. We find that going deeper does not improve performance. 
The best performing architectures are \glspl{MLP} with either 0 or 1 hidden layers. 
We use a tanh non-linearity as the activation function between layers. Note that for \mvae, since the output of $g_\phi(\cdot)$ is used as the mean and variance of a Gaussian random variable, we do not apply an activation function to it. Thus, the \mvae~ with 0-hidden-layer \gls{MLP} is effectively a log-linear model.
We tune the regularization parameter $\beta$ for \mvae~ following the procedure described in \Cref{sec:reg}. We anneal the Kullback-Leibler term linearly for 200,000 gradient updates. For both \mvae~ and \mdae, we apply dropout at the input layer with probability $0.5$. We apply a weight decay of $0.01$ for \mdae. We do not apply weight decay for any \gls{VAE} models.
We train both \mvae~ and \mdae~ using Adam \citep{kingma2014adam} with batch size of 500 users. For ML-20M, we train for 200 epochs. We train for 100 epochs on the other two datasets. We keep the model with the best validation NDCG@100 and report test set metrics with it.

\subsection{Baselines}
We compare results with the following standard state-of-the-art collaborative filtering models, both linear and non-linear:

\parhead{\Gls{WMF}} \citep{hu2008collaborative}: a linear low-rank factorization model. We train \gls{WMF}
with alternating least squares; this generally leads to
better performance than with SGD. We set the weights on all the 0's to 1 and tune the weights on all the 1's in the click matrix among $\{2, 5, 10, 30, 50, 100\}$, as well as the latent representation dimension $K \in \{100, 200\}$ by evaluating NDCG@100 on validation users.

\parhead{\textsc{Slim}} \citep{ning2011slim}: a linear model which
learns a sparse item-to-item similarity matrix by solving
a constrained $\ell_1$-regularized optimization problem. We grid-search both of the regularization parameters over $\{0.1, 0.5, 1, 5\}$ and report the setting with the best NDCG@100 on validation users. We did not evaluate \textsc{Slim} on MSD because the dataset is too large for it to finish in a reasonable amount of time (for the Netflix dataset, the parallelized grid search took about two weeks). We also found that the faster approximation of \textsc{Slim} \citep{levy2013efficient} did not yield competitive performance. 

\glsreset{CDAE}
\parhead{\Gls{CDAE}} \citep{wu2016collaborative}: augments the standard denoising autoencoder by adding a per-user latent factor to the input. We change the (user, item) entry subsampling strategy in SGD training in the original paper to the user-level subsampling as we did with \mvae~ and \mdae. We generally find that this leads to more stable convergence and better performance. We set the dimension of the bottleneck layer to 200, and use a weighted square loss, equivalent to what the square loss with negative sampling used in the original paper. We apply tanh activation at both the bottleneck layer as well as the output layer.\footnote{\citet{wu2016collaborative} used sigmoid activation function but mentioned tanh gave similar results. We use tanh to be consistent with our models.} We use Adam with a batch size of 500 users. As mentioned in \Cref{sec:related}, the number of parameters for \gls{CDAE} grows linearly with the number of users and items. Thus, it is crucial to control overfitting by applying weight decay. We select the weight decay parameter over $\{0.01, 0.1, \cdots, 100\}$ by examining the validation NDCG@100. 

\glsreset{NCF} 
\parhead{\Gls{NCF}} \citep{he2017neural}: explores non-linear interactions (via a neural network) between the user and item latent factors. Similar to \gls{CDAE}, the number of parameters for \gls{NCF} grows linearly with the number of users and items. We use the publicly available source code provided by the authors, yet cannot obtain competitive performance on the datasets used in this paper --- the validation metrics drop within the first few epochs over a wide range of regularization parameters. The authors kindly provided the two datasets (ML-1M and Pinterest) used in the original paper, as well as the training/test split, therefore we separately compare with \gls{NCF} on these two relatively smaller datasets in the empirical study. In particular, we compare with the hybrid NeuCF model which gives the best performance in \citet{he2017neural}, both with and without pre-training.

We also experiment with \gls{BPR} \citep{rendle2009bpr}. However, the performance is not on par with the other baselines above. This is consistent with some other studies with similar baselines \citep{sedhain2016effectiveness}. Therefore, we do not include \gls{BPR} in the following results and analysis. 

\subsection{Experimental results and analysis}
In this section, we quantitatively compare our proposed methods with various baselines. In addition, we aim to answer the following two questions:
\begin{enumerate}
	\item How does multinomial likelihood compare with other commonly used likelihood models for collaborative filtering?
	\item When does \mvae~ perform better/worse than \mdae?
\end{enumerate}

\begin{table}
\centering
\caption{Comparison between various baselines and our proposed methods. Standard errors
are around 0.002 for ML-20M and 0.001 for Netflix and MSD. Both \mvae~ and \mdae~ significantly outperform the baselines across datasets and metrics. We could not finish \textsc{Slim} within a reasonable amount of time on MSD.}
\begin{subtable}[t]{\columnwidth}
\caption{ML-20M}
	\centering
\begin{tabular}{ l c c c c }
    & Recall@20 & Recall@50 & NDCG@100  \\
  \toprule
  \mvae & \bf 0.395 & \bf 0.537 & \bf 0.426 \\
  \mdae & 0.387 & 0.524 & 0.419 \\
  \midrule 
  \gls{WMF} & 0.360 & 0.498 & 0.386 \\
  \textsc{Slim} & 0.370 & 0.495 & 0.401\\
  \gls{CDAE} & 0.391 & 0.523 & 0.418 \\
  \bottomrule\\
\end{tabular}
\end{subtable}

\begin{subtable}[t]{\columnwidth}
\caption{Netflix}
	\centering
\begin{tabular}{ l c c c c }
 & Recall@20 & Recall@50 & NDCG@100  \\
  \toprule
  \mvae & \bf 0.351 & \bf 0.444 & \bf 0.386\\
  \mdae & 0.344 & 0.438 & 0.380 \\
  \midrule 
  \gls{WMF} & 0.316 & 0.404 & 0.351\\
  \textsc{Slim} & 0.347 & 0.428 & 0.379 \\
  \gls{CDAE} & 0.343 & 0.428 & 0.376 \\
  \bottomrule\\
\end{tabular}
\end{subtable}

\begin{subtable}[t]{\columnwidth}
\caption{MSD}
	\centering
\begin{tabular}{ l c c c c }
 & Recall@20 & Recall@50 & NDCG@100  \\
  \toprule
  \mvae & \bf 0.266 & \bf 0.364 & \bf 0.316 \\
  \mdae  & \bf 0.266 & \bf 0.363 & 0.313 \\
  \midrule 
  \gls{WMF} & 0.211 & 0.312 & 0.257 \\
  \textsc{Slim} & --- & --- & --- \\
  \gls{CDAE} & 0.188 & 0.283 & 0.237\\
  \bottomrule\\
\end{tabular}
\end{subtable}
\label{tab:results}
\end{table}

\parhead{Quantitative results.} \Cref{tab:results} summarizes the results between our proposed methods and various baselines. Each metric is averaged across all test users. Both \mvae~ and \mdae~ significantly outperform the baselines across datasets and metrics. \mvae~ significantly outperforms \mdae~ on ML-20M and Netflix data-sets. In most of the cases, non-linear models (\mvae, \mdae, and \gls{CDAE}) prove to be more powerful collaborative filtering models than state-of-the-art linear models. The inferior results of \gls{CDAE} on MSD are possibly due to overfitting with the huge number of users and items, as validation metrics drop within the first few epochs even though the training objective continues improving.

\begin{table}
\centering
\caption{Comparison between \gls{NCF} and \mdae~ with $[I \rightarrow 200 \rightarrow I]$ architecture. We take the results of \gls{NCF} from \citet{he2017neural}. \mdae~ model significantly outperforms \gls{NCF} without pre-training on both datasets and further improves on Pinterest even comparing with pre-trained \gls{NCF}.}
\begin{subtable}[t]{\columnwidth}
\caption{ML-1M}
\centering
\begin{tabular}{ c c c c }
 & \gls{NCF} & \gls{NCF} (pre-train) & \mdae  \\
  \toprule
  Recall@10 & 0.705 & \bf 0.730 & 0.722 \\
  NDCG@10 & 0.426 & \bf 0.447 & 0.446 \\
  \bottomrule\\
\end{tabular}
\end{subtable}

\begin{subtable}[t]{\columnwidth}
\caption{Pinterest}
\centering
\begin{tabular}{ c c c c }
 & \gls{NCF} & \gls{NCF} (pre-train) & \mdae \\
  \toprule
  Recall@10 & 0.872 & 0.880 & \bf 0.886\\
  NDCG@10 & 0.551 & 0.558 & \bf 0.580  \\
  \bottomrule\\
\end{tabular}
\end{subtable}
\label{tab:ae_vs_ncf}
\end{table}

We compare with \gls{NCF} on the two relatively smaller datasets used in \citet{hu2008collaborative}: ML-1M (6,040 users, 3,704 items, 4.47\% density) and Pinterest (55,187 users, 9,916 items, 0.27\% density). Because of the size of these two datasets, we use \mdae~ with a 0-hidden-layer \gls{MLP} generative model --- the overall architecture is $[I \rightarrow 200 \rightarrow I]$. (Recall \mvae~ with a 0-hidden-layer \gls{MLP} generative model is effectively a log-linear model with limited modeling capacity.) \Cref{tab:ae_vs_ncf} summarizes the results between \mdae~ and \gls{NCF}. \mdae~ significantly outperforms \gls{NCF} without pre-training on both datasets. On the larger Pinterest dataset, \mdae~ even improves over the pre-trained \gls{NCF} model by a big margin. 

\parhead{How well does multinomial likelihood perform?} 
Despite being commonly used in language models, multinomial likelihoods have typically received less attention in the collaborative filtering literature, 
especially with latent-factor models. Most previous work builds on Gaussian likelihoods (square loss, \Cref{eq:gaussian}) \citep{hu2008collaborative,ning2011slim,wu2016collaborative} or logistic likelihood (log loss, \Cref{eq:logistic}) \citep{wu2016collaborative,he2017neural} instead. 
We argue in \Cref{sec:model} that multinomial likelihood is in fact a good proxy for the top-$N$ ranking loss and is well-suited for implicit feedback data. 
To demonstrate the effectiveness of multinomial likelihood, we take the best-performing \mvae~ and \mdae~ model on each dataset and swap the likelihood distribution model for the data while keeping everything else exactly the same. 

\begin{table}
\centering
\caption{Comparison of \mvae~ and \mdae~ with different likelihood functions at the output layer on ML-20M. The standard error is around 0.002 (the results on the other two datasets are similar.) The multinomial likelihood performs better than the other two commonly-used likelihoods from the collaborative filtering literature. }
\begin{tabular}{ l c c c c }
  & Recall@20 & Recall@50 & NDCG@100  \\
  \toprule
  \vae{Mult} & \bf 0.395 & \bf 0.537 & \bf 0.426 \\
  \vae{Gaussian}  & 0.383 & 0.523 & 0.415 \\
  \vae{Logistic}  & 0.388 & 0.523 & 0.419 \\
  \midrule
  \dae{Mult} & \bf 0.387 & \bf 0.524 & \bf 0.419 \\
  \dae{Gaussian} & 0.376 & 0.515 & 0.409 \\
  \dae{Logistic} & 0.381 & 0.516 & 0.414 \\
  \bottomrule
\end{tabular}
\label{tab:multi_vs_others}
\end{table}

\Cref{tab:multi_vs_others} summarizes the results of different likelihoods on ML-20M (the results on the other two datasets are similar.) We tune the hyperparameters for each likelihood separately.\footnote{Surprisingly, partial regularization seems less effective for Gaussian and logistic.} The multinomial likelihood performs better than the other likelihoods. The gap between logistic and multinomial likelihood is closer --- this is understandable since multinomial likelihood can be approximated by individual binary logistic likelihood, a strategy commonly adopted in language modeling \citep{mikolov2013distributed,xu2011efficient}.

We wish to emphasize that the choice of likelihood remains data-dependent. For the task of collaborative filtering, the multinomial likelihood achieves excellent empirical results.
The methodology behind the partial regularization in \mvae, however, is a technique we hypothesize will 
generalize to other domains. 

\begin{figure*}[!ht]
  \centering
  \centering
    \begin{subfigure}[b]{\textwidth}
  \includegraphics[width=\textwidth]{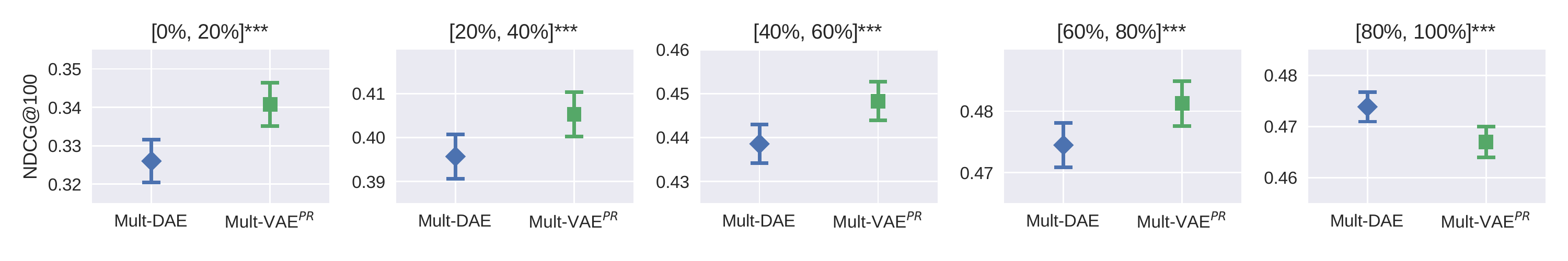}
        \caption{ML-20M}
        \label{fig:vae_ae_ml20m}
    \end{subfigure}

   \begin{subfigure}[b]{\textwidth}
  \includegraphics[width=\textwidth]{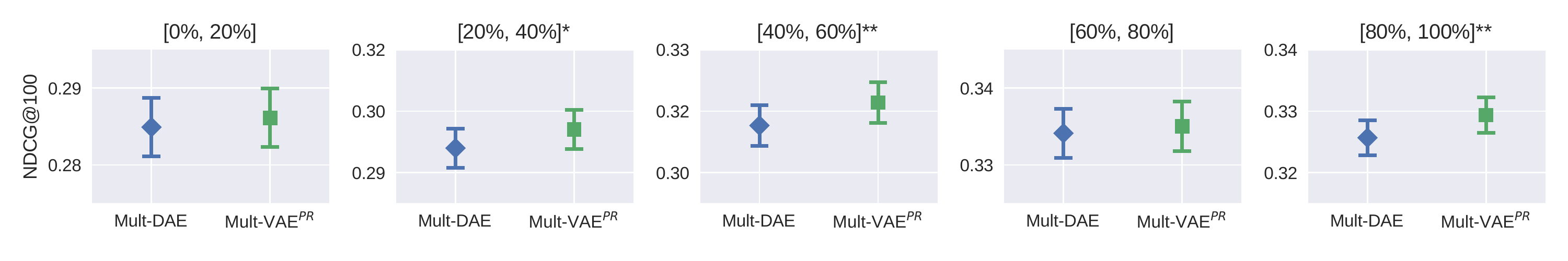}
        \caption{MSD}
        \label{fig:vae_ae_msd}
    \end{subfigure}

    \caption{NDCG@100 breakdown for users with increasing levels of activity (starting from $0\%$), measured by how many items a user clicked on in the fold-in set. The error bars represents one standard error. For each subplot, a paired t-test is performed and * indicates statistical significance at $\alpha=0.05$ level, ** at $\alpha=0.01$ level, and *** at $\alpha=0.001$ level. Although details vary across datasets, \mvae~ consistently improves recommendation performance for users who have only clicked on a small number of items.}\label{fig:vae_ae}

\end{figure*}

\parhead{When does \mvae~ perform better/worse than \mdae?}
In \Cref{tab:results} we can see that both \mvae~ and \mdae~ produce competitive empirical results with \mvae~ being comparably better. It is natural to wonder when a variational Bayesian inference approach (\mvae) will win over using a point estimate (\mdae) and vice versa. 

Intuitively, \mvae~ imposes stronger modeling assumptions and therefore could be more robust when user-item interaction data is scarce. To study this, we considered two datasets: ML-20M where \mvae~ has the largest margin over \mdae~ and MSD where \mvae~ and \mdae~ have roughly similar performance. The results on the Netflix dataset are similar to ML-20M. 
We break down test users into quintiles based on their activity level in the fold-in set which is provided as input to the inference model $g_\phi(\cdot)$ to make prediction. The activity level is simply the number of items each user has clicked on. We compute NDCG@100 for each group of users using both \mvae~ and \mdae~ and plot results in \Cref{fig:vae_ae}. This summarizes how performance differs across users with various levels of activity. 

In \Cref{fig:vae_ae}, we show performance across increasing user activity. Error bars represents one standard error. For each subplot, a paired t-test is performed and statistical significance is marked. Although there are slight variations across
datasets, \mvae~ consistently improves recommendation performance for users who have only clicked on a small number of items. This is particularly prominent for ML-20M (\Cref{fig:vae_ae_ml20m}). Interestingly, \mdae~ actually outperforms \mvae~ on the most active users. This indicates the stronger prior assumption could 
potentially hurt the performance when a lot of data is available for a user. 
For MSD (\Cref{fig:vae_ae_msd}), the least-active users have similar performance under both \mvae~and \mdae. However, as we described in \Cref{sec:exp_data}, MSD is pre-processed so that a user has at least listened to 20 songs. Meanwhile for ML-20M, each user has to watch at least 5 movies. This means that the first bin of ML-20M has much lower user activity than the first bin of MSD.  

Overall, we find that \mvae, which may be viewed under the lens of a principled Bayesian inference approach, is more robust than the point estimate approach of \mdae, regardless of the scarcity of the data. More importantly, the \mvae~ is less sensitive to the choice of hyperparameters -- weight decay is important for \mdae~ to achieve competitive performance, yet it is not required for \mvae.
On the other hand, \mdae~ also has advantages: it requires fewer parameters in the bottleneck layer --- \mvae~ requires two sets of parameters to obtain the latent representation $\mbz$: one set for the variational mean $\mu_\phi(\cdot)$ and another for the variational variance $\sigma_\phi(\cdot)$ --- and \mdae~ is conceptually simpler for practitioners. 
 
\section{Conclusion}

In this paper, we develop a variant of \gls{VAE} for collaborative filtering on implicit feedback data. This enables us to go beyond linear factor models with limited modeling capacity. 

We introduce a generative model with a multinomial likelihood function parameterized by neural network. We show that multinomial likelihood is particularly well suited to modeling user-item implicit feedback data. 

Based on an alternative interpretation of the \gls{VAE} objective, we introduce an additional regularization parameter to partially regularize a \gls{VAE} (\mvae). We also provide a practical and efficient way to tune the additional parameter introduced using KL annealing. We compare the results obtained against a denoising autoencoder (\mdae). 

Empirically, we show that the both \mvae~and \mdae~provide competitive performance with \mvae~ significantly outperforming the state-of-the-art baselines on several real-world datasets, including two recently proposed neural-network-based approaches. Finally, we identify the pros and cons of both \mvae~ and \mdae~ and show that employing a principled Bayesian approach is more robust.

In future work, we would like to futher investigate the trade-off introduced by the additional regularization parameter $\beta$ and gain more theoretical insight into why it works so well. 
Extending \mvae~by \emph{condition} on side information might also be a way to improve performance.

\end{document}